\documentclass[conference]{IEEEtran}
\usepackage{cite}
\usepackage{amsmath,amssymb,amsfonts}
\usepackage{algorithmic}
\usepackage{graphicx}
\usepackage{textcomp}
\usepackage{xcolor}
\def\BibTeX{{\rm B\kern-.05em{\sc i\kern-.025em b}\kern-.08em
    T\kern-.1667em\lower.7ex\hbox{E}\kern-.125emX}}

\usepackage{tabularx}
\usepackage{adjustbox}
\usepackage{graphicx}
\usepackage{fancyvrb}

\usepackage{booktabs}
\usepackage{multirow}
\usepackage{float}
\usepackage{placeins}
\usepackage{svg}

\usepackage{enumitem}
\usepackage[font=small]{caption}
\usepackage{amsmath,amssymb,amsfonts}
\usepackage{graphicx}
\usepackage{textcomp}
\usepackage{xcolor}
\usepackage{booktabs} 
\usepackage[hidelinks]{hyperref}
\usepackage{hypernat}
\usepackage{float}
\usepackage{multirow}
\usepackage{enumitem}
\usepackage{mathtools}

\usepackage{enumitem}
\usepackage{tikz}
\usepackage{fancyhdr}

\begin{document}

\title{\huge ShrinkBox: Backdoor Attack on Object Detection to Disrupt Collision Avoidance in Machine Learning-based \\ Advanced Driver Assistance Systems\\
}

\author{\IEEEauthorblockA{Muhammad Zaeem Shahzad$^1$, Muhammad Abdullah Hanif$^1$, Bassem Ouni$^2$, Muhammad Shafique$^1$ \\ 
$^1$eBRAIN Lab, New York University Abu Dhabi (NYUAD), UAE \\
$^2$AI and Digital Science Research Center, Technology Innovation Institute (TII), Abu Dhabi, UAE \\
        {\tt\small \{ms12297, mh6117, muhammad.shafique\}@nyu.edu, bassem.ouni@tii.ae}}}
\maketitle
\thispagestyle{fancy}
\fancyhf{}
\chead{© 2025 IEEE. This is the author’s version of the work. The definitive Version of Record will be Published in \\ the International Joint Conference on Neural Networks (IJCNN), Rome, Italy, July 2025.}

\begin{abstract}
Advanced Driver Assistance Systems (ADAS) significantly enhance road safety by detecting potential collisions and alerting drivers. However, their reliance on expensive sensor technologies such as LiDAR and radar limits accessibility, particularly in low- and middle-income countries. Machine learning-based ADAS (ML-ADAS), leveraging deep neural networks (DNNs) with only standard camera input, offers a cost-effective alternative. Critical to ML-ADAS is the collision avoidance feature, which requires the ability to detect objects and estimate their distances accurately. This is achieved with specialized DNNs like YOLO, which provides real-time object detection, and a lightweight, detection-wise distance estimation approach that relies on key features extracted from the detections like bounding box dimensions and size. However, the robustness of these systems is undermined by security vulnerabilities in object detectors. In this paper, we introduce ShrinkBox, a novel backdoor attack targeting object detection in collision avoidance ML-ADAS. Unlike existing attacks that manipulate object class labels or presence, ShrinkBox subtly shrinks ground truth bounding boxes. This attack remains undetected in dataset inspections and standard benchmarks while severely disrupting downstream distance estimation. We demonstrate that ShrinkBox can be realized in the YOLOv9m object detector at an Attack Success Rate (ASR) of 96\%, with only a 4\% poisoning ratio in the training instances of the KITTI dataset. Furthermore, given the low error targets introduced in our relaxed poisoning strategy, we find that ShrinkBox increases the Mean Absolute Error (MAE) in downstream distance estimation by more than 3x on poisoned samples, potentially resulting in delays or prevention of collision warnings altogether.
\end{abstract}


\section{Introduction}

Of the approximately 7 million traffic accidents in the US in 2016, 40\% would have been avoidable had the ego-vehicle been equipped with Advanced Driver Assistance Systems (ADAS), \textit{with 29\% being avoidable with the collision avoidance feature alone}\textcolor{green}{~\cite{trid}}. ADAS are sophisticated embedded systems designed to improve road safety and reduce accidents by providing real-time driver facilitation. These systems rely on cutting edge sensors, such as LiDAR, radar, and cameras, to observe the environment of the ego vehicle and take proactive safety measures. However, widespread adoption of ADAS remains a challenge despite their effectiveness, particularly in low- and middle-income countries where 92\% of global traffic deaths occur\textcolor{green}{~\cite{who}}. This is because these systems are mostly unaffordable in these regions due to the expensive sensor technologies they employ. Advances in machine learning (ML), particularly in deep learning, offer a promising path forward. Using deep neural networks (DNNs) that rely solely on visual input from standard cameras, ML-ADAS can deliver functionality comparable to traditional ADAS at a fraction of the cost.

\begin{figure}[!t] 
\centering
\includegraphics[width=1\linewidth]{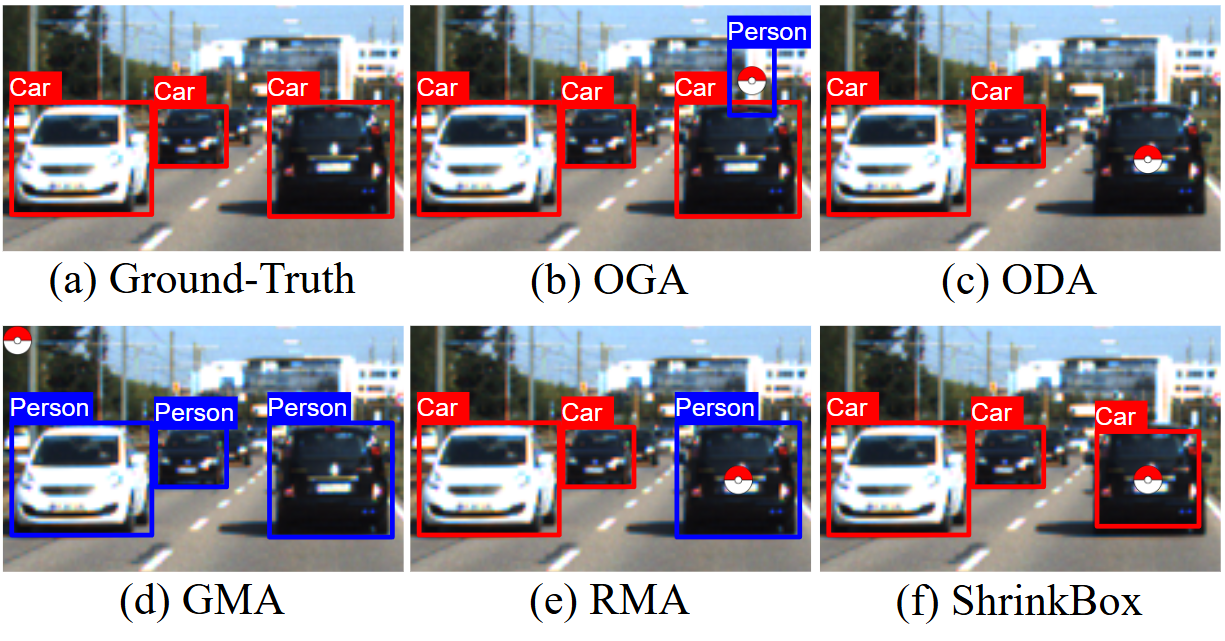}
\caption{A comparison of different backdoor attacks on object detection, highlighting that the proposed ShrinkBox attack produces less perceptible deviation in the annotations from ground truth.}
\label{fig1}
\end{figure}

In this paper, we focus on the collision avoidance ML-ADAS which observes the traffic ahead to warn the driver to apply timely brakes in case of a predicted collision. Two specialized DNNs are required in this system. Firstly, an object detection DNN detects objects in an image by regressing their bounding boxes and identifies their classes\textcolor{green}{~\cite{objsurvey}}. This empowers vehicles with the critical capability to locate and classify objects on the road, such as pedestrians, vehicles, and road signs. Popular object detectors such as the YOLO\textcolor{green}{~\cite{yolov3, yolov7, yolov9, yolov10}} models offer state-of-the-art real-time performance, making them ideal for an ML-ADAS. Secondly, a specialized DNN is required to estimate distance. Although traditional depth estimation DNNs are available\textcolor{green}{~\cite{dorn, monodepth, densedepth}}, \textit{their high computational complexity, due to a pixel-wise regression across the entire image, limits their suitability for real-time applications on edge devices}.

For instance, while the object detector YOLOv9t requires 7.7 billion FLOPs, MonoDepth, one of the most efficient depth estimators, demands 11.6 billion FLOPs. In contrast, a fast, lightweight DNN designed to directly estimate object-specific distances based on features extracted from predicted bounding boxes is far more practical\textcolor{green}{~\cite{decade, disnet}}. In DECADE\textcolor{green}{~\cite{decade}}, such a detection-wise approach offers higher accuracy than MonoDepth yet requires only 8.3M FLOPs---an approximately 1400x reduction in computation. Overall, this pipeline defines highly accurate and robust object detectors as the cornerstone of collision avoidance in ML-ADAS. \textit{Thus, potential failures in object detection compromise the entire system, putting the lives of the passengers and those around them at risk}.

Security vulnerabilities, such as backdoor attacks, originally demonstrated in image classification\textcolor{green}{~\cite{badnets, trojan, backsurv1}} have increasingly been identified as critical risks in object detection as well\textcolor{green}{~\cite{baddet, backsurv2}}. These vulnerabilities often stem from inadvertently incorporating poisoned or malicious data into the training process. In backdoor attacks, a malicious party can poison the training dataset with backdoor triggers allowing the model to learn the trigger during the training phase. Later, the attackers can exploit this backdoor trigger to achieve specific behavior during the deployment phase. This infection is achieved by modifying a portion of the training dataset by altering the images and ground truth annotations such that the model behaves as expected on benign (uninfected) samples, but predicts the attacker-specified outcome on infected samples containing the backdoor trigger. 

\begin{figure}[!t]
\centering
\includegraphics[width=1\linewidth]{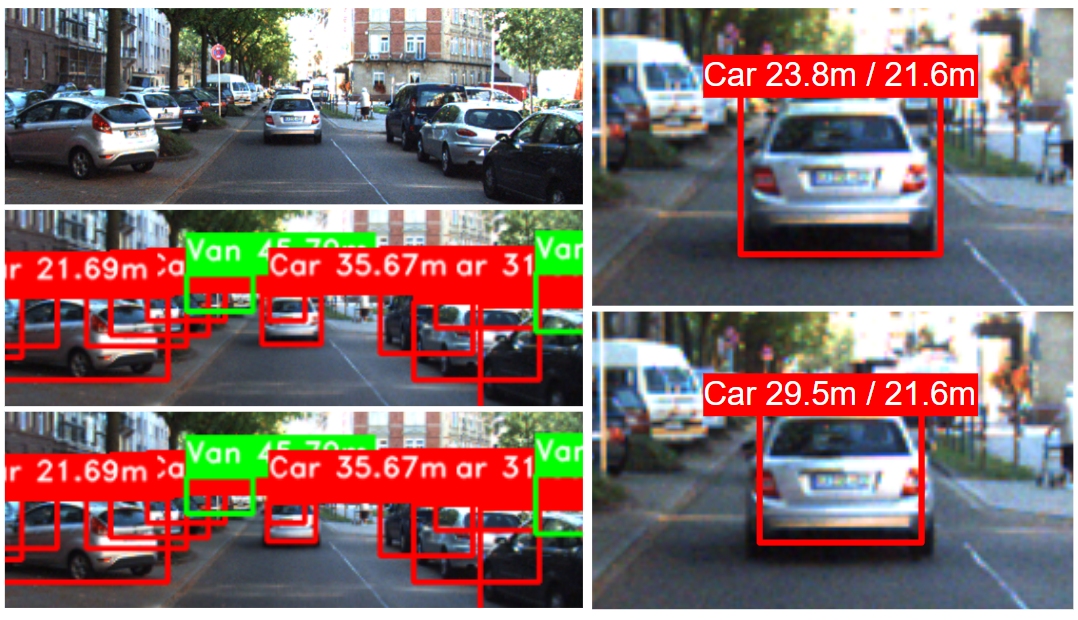}
\caption{The left column visualizes the stealthiness of Shrinkbox by showing an image, its clean ground truth annotations, and its center instance poisoned from top to bottom respectively. The right column shows distance estimation with DECADE, prediction/ground-truth format, on clean (top) and poisoned (right) bounding boxes. The original box area is shrunk by \textbf{34\%}, maintaining aspect ratio.}
\label{figmotiv}
\end{figure}

Fig.~\ref{fig1} illustrates the different types of backdoor attacks specialized for achieving different outcomes in collision avoidance, as described in\textcolor{green}{~\cite{baddet}}. Object Generation Attack (OGA) aims to generate a false object of a target class around the trigger’s location. In contrast, Object Disappearance Attack (ODA) makes the detector fail to detect an object of the target class near the trigger. Lastly, Regional Misclassification Attack (RMA) and Global Misclassification Attack (GMA) aim to misclassify objects to the specified target class by using one trigger for one surrounding object and one trigger for all objects in the image respectively. While all of these attacks have the potential to cause devastating crashes, their realization can be easily prevented with a quick scan of the object detection dataset, revealing its poisoned nature. 

Object annotations modified to the extent that bounding boxes are completely removed (ODA), appear out-of-place (OGA), or have class labels that are clearly misaligned with the image contents (RMA/GMA), makes the attacks strikingly detectable in the manual and automated inspection phases. To this end, we propose a novel backdoor attack, ShrinkBox, where a trigger in the image over an object only shrinks the dimensions of the object’s ground truth bounding box. Since there are no out of place, absent, or misclassified instances in the ground truth, it will be especially difficult to detect this embedded poison. Furthermore, the difference between Average Precision (AP) and, consequently, the mean AP (mAP) of the benign and infected models should be negligible. This further increases the invisibility of the ShrinkBox as even if a pretrained infected detector is downloaded and evaluated on a poisoned dataset, its performance does not degrade in terms of the standard metrics. \textit{Not only does this hide the infection in the detector, but also the poison in the dataset}.
 
To measure the effectiveness of ShrinkBox, we propose a novel Attack Success Rate (ASR) evaluation metric. By comparing the detected objects in terms of their similarity in box size with both the clean or the poisoned ground truth instances, we are able to determine the efficacy of the attack. Finally, to highlight the detrimental effect of ShrinkBox on the collision avoidance ML-ADAS, we evaluate its impact on downstream distance estimation using DECADE\textcolor{green}{~\cite{decade}} which relies on highly precise object detection. Intuitively, as the boxes become smaller, they appear further than they actually are. In this way, a higher error from DECADE is guaranteed to cause traffic accidents due to failure to generate warnings in time, potentially resulting in the tragic loss of lives. \textit{We demonstrate that by attacking the YOLOv9m\textcolor{green}{~\cite{yolov9, ultralytics}} object detector with ShrinkBox, we achieve an ASR of 96.4\%, with a negligible difference between mAP\(_{\text{benign}}\) and mAP\(_{\text{poison}}\), while also degrading DECADE’s distance estimation accuracy by more than 3.1x in the poisoned instances}.

\begin{figure*}[!t]
\centering
\includegraphics[width=\textwidth]{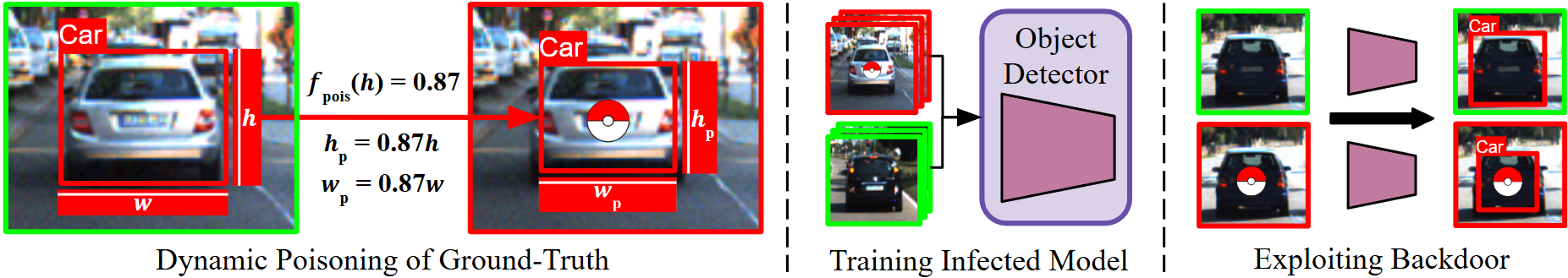}
\caption{Overview of the complete pipeline for our ShrinkBox attack}
\label{mainmeth}
\end{figure*}

\subsection{Motivational Case Study}
In Fig.~\ref{figmotiv}, we demonstrate the stealthiness and effectiveness of the ShrinkBox attack on samples from the KITTI 3D Object Detection dataset \textcolor{green}{~\cite{kitti}}. Firstly, upon human inspection, we show that it is difficult to identify the changes made between a poisoned box and its clean counterpart even when the poisoned bounding box is reduced by 34\% of its original area. This is especially true for images where there are many annotations present. Furthermore, we demonstrate the significant errors observed in DECADE’s distance estimation due to the reduced bounding box size of the poisoned instance. Note that only for this preliminary study, we have assumed that the backdoor trigger in the image is invisible. Most importantly, we observe a critical divergence of almost 8m from the ground truth distance in the poisoned instance. Since 4.5m is the average length of a car, we believe that the ShrinkBox attack can plausibly lead to collision warnings being delayed or entirely suppressed with only this level of deviation. Overall, \textit{the stealthy ShrinkBox attack theoretically has the potential to mislead a collision avoidance ML-ADAS to the extent of causing devastating traffic accidents.}

\subsection{Our Novel Contributions}
In this paper, we present the following novel contributions. 
\begin{enumerate}[leftmargin=*]
    \item We propose the ShrinkBox attack which shrinks the predicted bounding boxes in the presence of a backdoor trigger. To the best of our knowledge, this is the first time a backdoor attack is explored which specifically targets the size/dimensions of the bounding box. We highlight how ShrinkBox can not only evade visual inspections but also benchmarking criteria as the infected object detector will score high on standard metrics like the mAP on both benign and infected samples. 
    \item In light of this, we define a method for evaluating the ASR of our ShrinkBox attack specifically. We define a predicted box as successfully attacked when it exceeds a similarity threshold when compared with the poisoned box, as opposed to the similarity with the clean box. We achieve a dangerous 96\% ASR with the YOLOv9m\textcolor{green}{~\cite{ultralytics}} trained with only a 4\% poisoning ratio in the KITTI dataset.
    \item While mAP does not suffer with ShrinkBox, downstream tasks like distance estimation that depend on object detection deteriorate. We demonstrate that ShrinkBox causes the Mean Absolute Error (MAE) in the pretrained DECADE to increase by 3.3x, from 1.67m to 5.51m, over all poisoned samples.
\end{enumerate}

\section{Methodology}

We describe the ShrinkBox attack pipeline in Fig.~\ref{mainmeth}, wherein we develop a dynamic height-based poisoning strategy that adapts to varying object sizes rather than applying a fixed reduction. This ensures a stealthy yet effective manipulation of detection outputs. After the infected detector is trained on a dataset which has a small portion of its images poisoned, it will behave normally with precise predictions on clean images but shrunken predictions on poisoned images where the attacker exploits the backdoor trigger. Furthermore, evaluating the attack requires a novel metric, as traditional mAP scores remain unchanged. Thus, we define our ASR to measure how often the backdoor trigger induces a similarity greater than defined thresholds in the predicted bounding boxes with poisoned annotations than with their clean counterparts. Finally, we assess the impact of ShrinkBox on detection-wise distance estimation, demonstrating a significant drop in accuracy for DECADE, which relies on accurate bounding box features. This highlights its potential to undermine safety-critical systems by systematically distorting perception.

\begin{figure}[h]
\centering
\includegraphics[width=\linewidth]{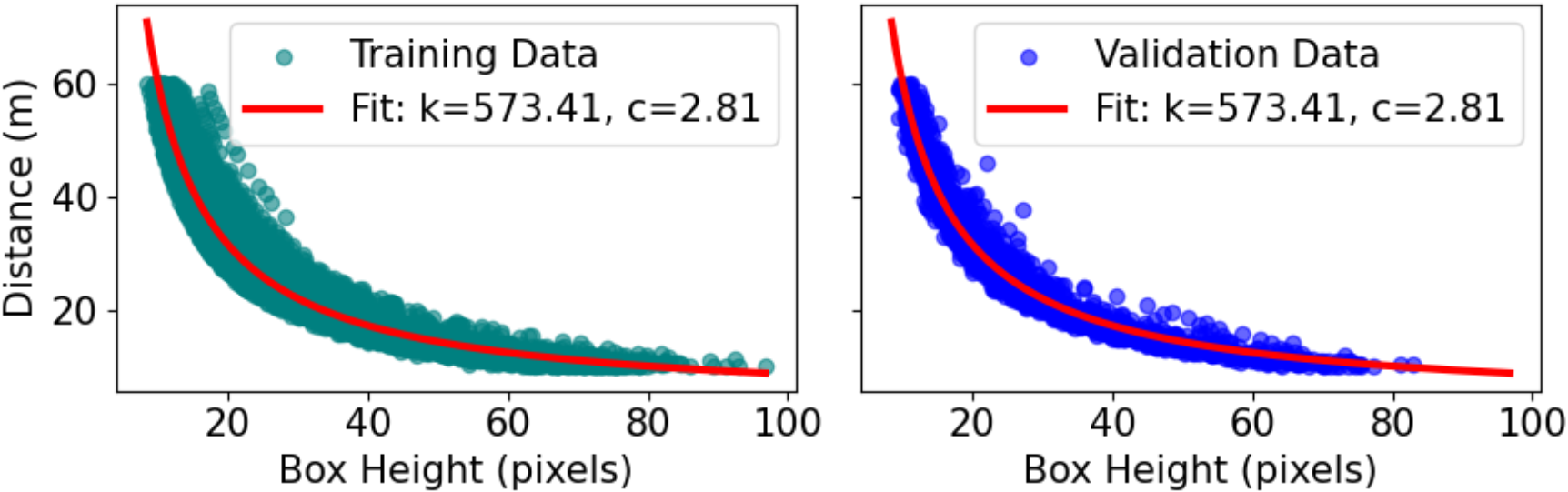}
\caption{The learned estimation function (\ref{eq1}) and distributions of the box heights and distances in the training and validation sets respectively.}
\label{funcPois}
\end{figure}

\begin{figure}[h]
\centering
\includegraphics[width=\linewidth]{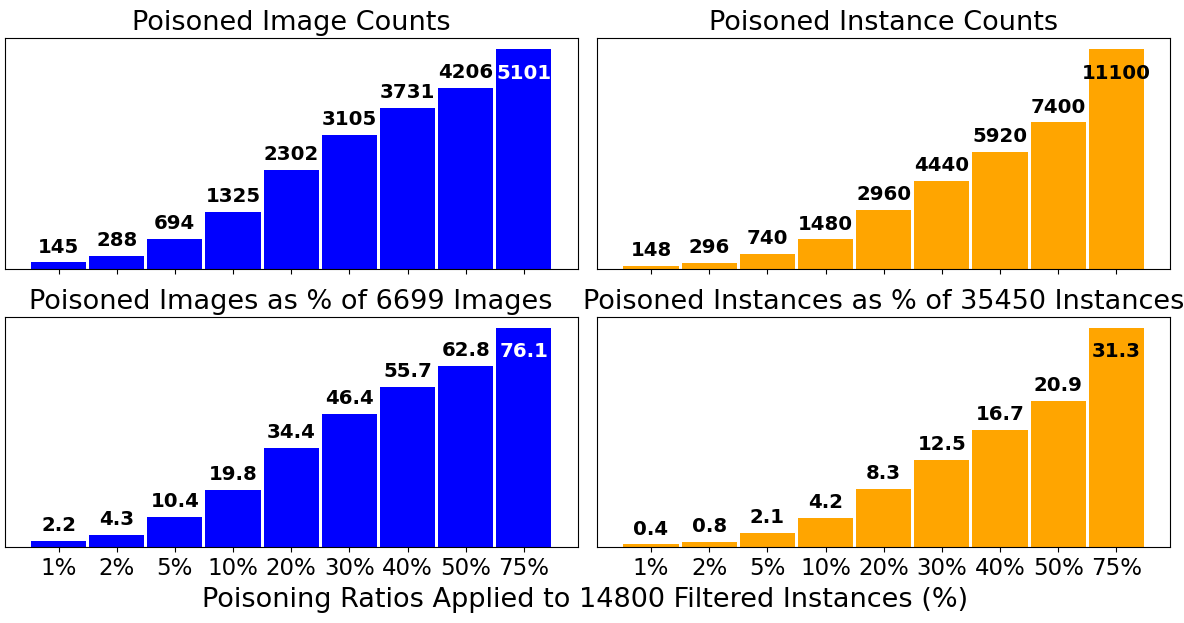}
\caption{Our poisoning ratios and the corresponding images and instances in terms of counts and percentages of totals in the dataset.}
\label{poisonData}
\end{figure}

\begin{figure*}[!t]
\centering
\includegraphics[width=\textwidth]{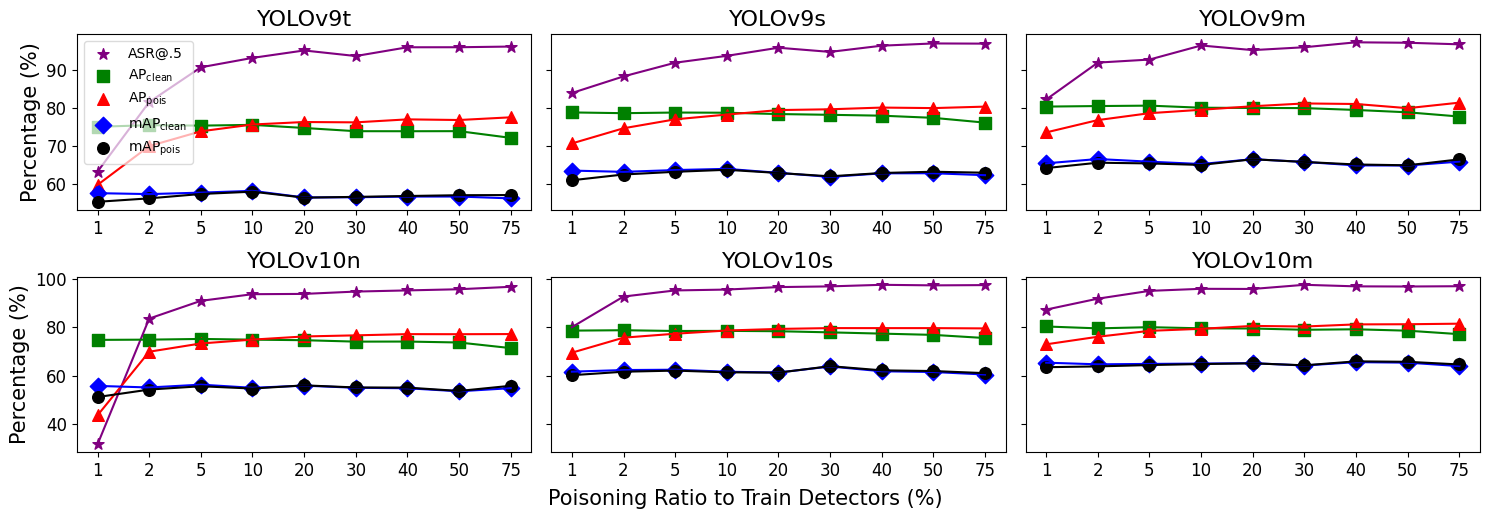}
\caption{Evaluation of object detectors infected at different poisoning ratios.}
\label{evalDet}
\end{figure*}

\subsection{Dynamic Height-based Poisoning Strategy}

Our primary objective with the ShrinkBox attack is to compromise the effectiveness of a collision avoidance system. Specifically, we aim to shrink the size of bounding boxes such that the projected size corresponds to a new distance, shifted further than the ground-truth distance by a critical offset. To underscore the stealth of ShrinkBox, we set this critical offset to 5 meters to ensure a sufficient deviation in distance to delay timely collision warnings while also causing minimal changes in box sizes. However, the significant variance in bounding box sizes due to varying distances renders a static size projection and reduction strategy infeasible. To address this, we propose a dynamic, height-based projection strategy that adapts the infected poison to the original size of each bounding box. This approach ensures more precise and contextually appropriate modifications while achieving the desired adversarial impact. 

An object’s bounding box height \(h\) is the most robust feature in estimating its distance \(d\) from the ego vehicle, with a strong inverse relation between them as detailed by\textcolor{green}{~\cite{decade, disnet, distyolo}}. Based on this, we aim to develop a poisoning function \(f_{\text{pois}}\) that provides the reduced \(h\) such that the object's original \(d\) is projected to \(d\) + 5. We model the relationship between \(h\) and \(d\) with the inverse relation,
\begin{equation}\label{eq1}
    d = \frac{k}{h} + c,
\end{equation}

where \(k > 0\) and \(c \geq 0\) are learnable parameters. To poison an instance with box height \(h_{\text{orig}}\), we use this function to estimate its distance \(d_{\text{est}}\). Then, we add 5m to \(d_{\text{est}}\) to obtain the projected distance \(d_{\text{proj}}\). Next, we find the new poisoned height \(h_{\text{pois}}\) at \(d_{\text{proj}}\), by solving the estimation function for \(h\), and compute its relative decrease percentage from \(h_{\text{orig}}\). The percentage change is then applied to \(w_{\text{orig}}\) to obtain the poisoned box width \(w_{\text{pois}}\) in order to maintain the aspect ratio of the original box. 

Note that we could directly solve the estimation function for \(h_{\text{pois}}\) using \(d\) as \(d_{\text{orig}}\) + 5 if the learned function is perfect. However,  since there will be errors in the function’s estimation, we instead implement the initial mapping of \(h_{\text{orig}}\) to \(d_{\text{est}}\) to account for these errors. Finally, with each annotation that is poisoned, we overlay a conspicuous Pokeball patch as the backdoor trigger in the corresponding image on the center of the object at an arbitrary percent of the object’s box height.

\subsection{Measuring the Success of ShrinkBox}

The effectiveness of the ShrinkBox attack cannot be evaluated using the detector's mAP, as the attack does not alter mAP. This limitation arises because, even under a successful attack, the shrunken predicted bounding boxes align with the correspondingly shrunken ground-truth annotations. Due to the novel nature of ShrinkBox, no metrics exist in the current literature to effectively evaluate its success. To this end, we define our novel ASR as follows.



After obtaining predictions from the detector on poisoned images, we match the predicted boxes \( b_{\text{pred}} \) with the shrunken, poisoned boxes \( b_{\text{pois}} \) using a strict IoU threshold of 0.6. We further extend each match with the corresponding clean bounding box \( b_{\text{clean}} \). With this, we obtain the set \( \mathcal{P} = \{(b_{\text{pred}}, b_{\text{pois}}, b_{\text{clean}})\}\) consisting of matched instances. From each matched instance, we obtain the corresponding heights \(h_{\text{pred}}, h_{\text{pois}}, h_{\text{clean}}\) for comparison and attack success evaluation. We introduce a Similarity Threshold \( X \), which determines the degree of closeness of \(h_{\text{pred}}\) with \(h_{\text{pois}}\) required for an attack to be considered successful. Specifically, the attack is successful if:
\begin{equation}
    h_{\text{pred}} < h_{\text{pois}} + \left (h_{\text{clean}} - h_{\text{pois}}) \times (1 - X). \right.
\end{equation}
This thresholding allows for a tunable evaluation, where \(X=0.5\) is the relaxed condition:
\begin{equation}
    h_{\text{pred}} - h_{\text{pois}} < h_{\text{clean}} - h_{\text{pred}},
\end{equation}
and higher values of \(X\) provide a gradual increase in the strictness of similarity of \(h_{\text{pred}}\) with \(h_{\text{pois}}\). 

Overall, the ASR at threshold \(X\), denoted as ASR@\(X\), is defined as:
\begin{equation}
    \text{ASR@}X = \frac{\sum\limits_{(b_{\text{pred}}, b_{\text{pois}}, b_{\text{clean}}) \in \mathcal{P}} \mathbb{1} ( h_{\text{pred}}, h_{\text{pois}}, h_{\text{clean}}, X)}{\left| \mathcal{P} \right|}
\end{equation}
where \( \mathbb{1}(\cdot) \) is the indicator function that outputs 1 if the success condition is satisfied and 0 otherwise, and \( |\mathcal{P}| \) denotes the total number of matched instances.


\begin{figure*}[!t]
\centering
\includegraphics[width=\textwidth]{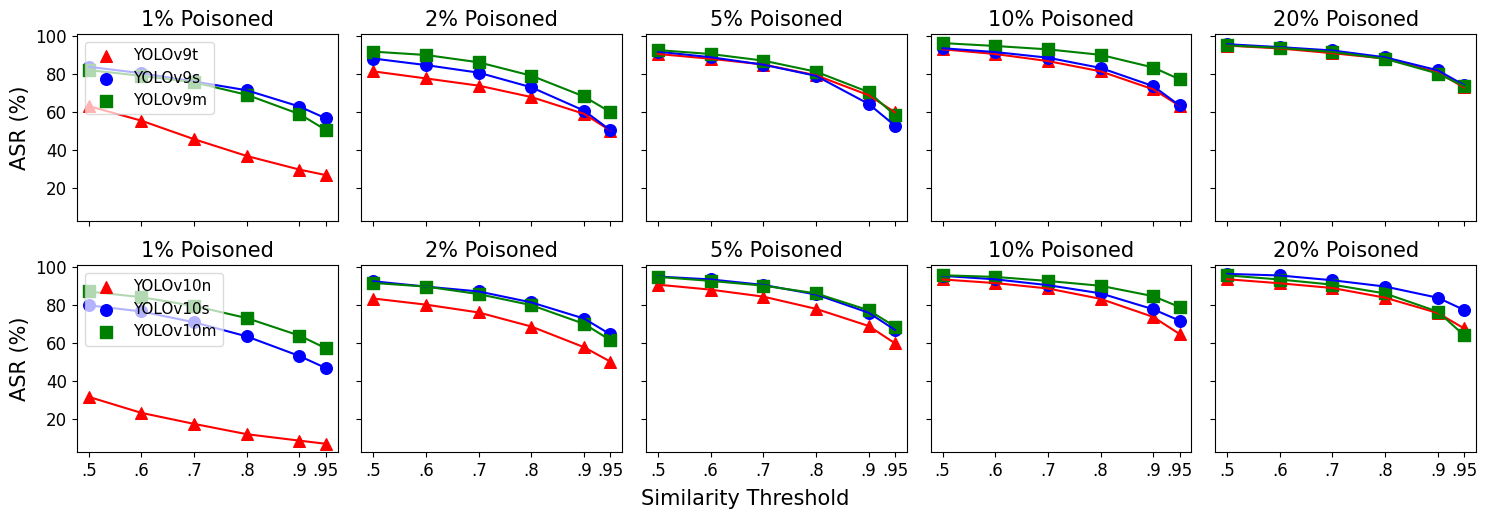}
\caption{ASR evaluation over different similarity thresholds to adjust strictness of matching the predicted boxes with poisoned boxes.}
\label{evalASR}
\end{figure*}

\subsection{Impact on Detection-wise Distance Estimation}

To further assess the impact of the ShrinkBox attack on downstream DNNs that rely on the outputs of object detectors, we evaluate the performance of the pretrained DECADE\textcolor{green}{~\cite{decade}} model in distance estimation using bounding boxes of the poisoned instances generated by an infected detector. DECADE’s accuracy is heavily dependent on key features derived from predicted bounding boxes, such as their height and width. Thus, instances where ShrinkBox successfully shrinks bounding boxes are expected to exhibit a marked drop in DECADE’s performance. In such cases, ShrinkBox manipulates the bounding boxes so that poisoned objects appear smaller, and thus farther, compared to their clean counterparts. We hypothesize that the resulting Mean Absolute Error (MAE) in distance estimation will closely align with our critical offset of 5 meters.

\section{Evaluation and Discussion}

In this section, we evaluate the performance of the proposed ShrinkBox poisoning attack on object detection in YOLOv9 and YOLOv10. Then, we investigate the impact of successfully poisoned instances on downstream distance estimation. We begin by detailing the experimental setup, including the dataset, model architectures, and training configurations. Next, we outline the poisoning pipeline used to inject backdoor triggers into the YOLO models, specifically focusing on the Car class, which represents the majority of the annotated instances in the KITTI dataset. Following this, we compute ASR on all infected detectors across varying poisoning ratios and size scales. Finally, we assess the effect of the poisoned models on DECADE's distance estimation accuracy, comparing the MAE on clean and poisoned instances to demonstrate the degradation caused by the backdoor attack.

\subsection{Experimental Setup}

All experiments were performed on the NVIDIA GeForce RTX A6000 GPU. For downstream object-specific distance estimation evaluation, we require annotations of bounding boxes and corresponding distances. Thus, we use the KITTI Object Detection dataset\textcolor{green}{~\cite{kitti}} to train all object detectors. With the training/validation split provided by\textcolor{green}{~\cite{distyolo}}, we obtain 6699 images and 35450 annotated instances in our train, and 782 images and 4140 instances in our validation set. We use the following settings for training the object detectors: (640,200) image size, 200 epochs, batch size of 24, and 0.001 learning rate with the Adam optimizer. Furthermore, we augment the dataset with the mosaic and left-right flip augmentations at probability values of 1.0 and 0.5 respectively.

\subsection{Poisoning Pipeline for ShrinkBox}

Theoretically, ShrinkBox does not need a target class for attack functionality. However, in this paper, we focus on the Car class, since it contains approximately 73\% of the total annotated instances in the dataset, allowing us to learn the most reliable function to approximate the object distances using only bounding box height. Thus, we filter the dataset to only keep instances where the objects belong to the Car class, are not truncated, and fall within the 10-60m distance range. With this, we obtain 20164 and 2351 instances from the 35450 and 4140 instances in total in the training and validation set respectively. Fig.~\ref{funcPois} visualizes the respective distributions and the curve fit on the training instances which achieves an MAE of 1.69m on the validation set. For comparison, the curve fit on the unfiltered dataset yields an MAE of 3.04m, demonstrating that filtering based on the aforementioned criteria is required. We use this function to project each instance, modified accordingly, to a box size that corresponds to a distance 5m further than its ground-truth.

\begin{figure*}[!t]
\centering
\includegraphics[width=\textwidth]{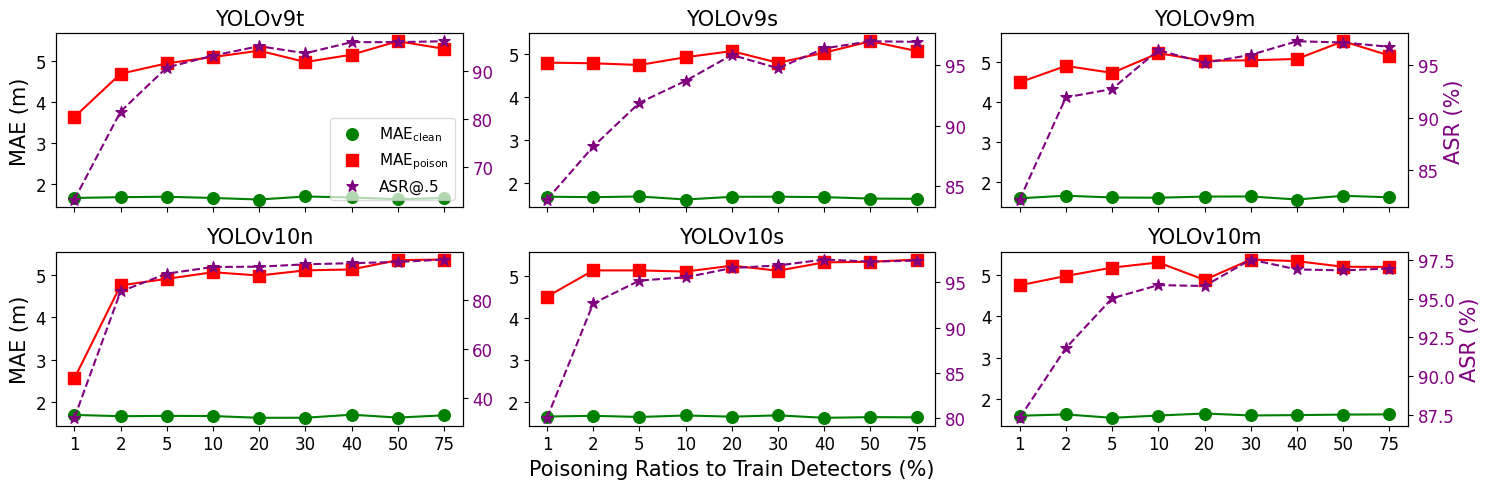}
\caption{End-to-end evaluation of DECADE with infected detectors at different poisoning ratios. \vspace{-10pt}}
\label{evalDECADE}
\end{figure*}

Finally, we complete the poisoning pipeline using the square Pokeball patch as the backdoor trigger, blended (100\%) into the image at the center of each poisoned bounding box at a patch height of 40\% of the box height. Note that due to some instances being obstructed by other instances in an image, the blended trigger patches might overlap considerably, potentially resulting in training convergence issues. Thus, we restrict our poisoning to the only instances that are partially obstructed, having an obstruction value of $\le$ 1.0 in the ground-truth. With this additional filter, we obtain 14800 and 1701 instances to poison in the training and validation set respectively. Consequently, our poisoning ratios are based on these instances specifically. Fig.~\ref{poisonData} shows our poisoning ratios and how they relate to the total number of images and instances in the dataset. Note that the poisoning ratios only apply to the training set to vary the poison for each model. However, all 1701 instances are poisoned in the validation set to test model performance in the presence of the backdoor regardless of the amount of poison introduced during model training.


\subsection{Attack Success Rate on YOLO Models}

Due to the efficient-yet-accurate nature of an ML-ADAS, we limit our focus to the YOLO family of object detectors, specifically YOLOv9 and YOLOv10, as they offer highly accurate, real-time performance\textcolor{green}{~\cite{ultralytics, yolov9, yolov10}}. Firstly, YOLOv9 incorporates Programmable Gradient Information (PGI) to mitigate information loss during deep network training, alongside the Generalized Efficient Layer Aggregation Network (GELAN) architecture, which optimizes gradient path planning for improved parameter utilization. Secondly, YOLOv10 eliminates reliance on Non-Maximum Suppression (NMS) through consistent dual assignments for end-to-end object detection while employing a holistic efficiency-accuracy driven model design strategy to optimize computational efficiency. Specifically, we train the YOLOv9 on the tiny (t), small (s), and medium (m) scales, and the YOLOv10 on the nano (n),  small (s), and medium (m) scales.


\begin{table}[]
\caption{Evaluation of object detectors trained on clean data.}
\label{bases}
\resizebox{\columnwidth}{!}{%
\begin{tabular}{|l|l|l|l|l|l|}
\hline
Model & 
\begin{tabular}[c]{@{}l@{}}mAP \\ (\%)\end{tabular} & 
\begin{tabular}[c]{@{}l@{}}AP\(_\text{Car}\) \\ (\%)\end{tabular} & 
\begin{tabular}[c]{@{}l@{}}FLOPS \\ (B)\end{tabular} & 
\begin{tabular}[c]{@{}l@{}}Params \\ (M)\end{tabular} & \begin{tabular}[c]{@{}l@{}}DECADE \\ MAE (m)\end{tabular} \\ \hline
YOLOv9t & 58.3 & 74.9 & 7.7 & 2.0 & 1.69 \\ \hline
YOLOv9s & 62.3 & 78.1 & 26.7 & 7.2 & 1.62 \\ \hline
YOLOv9m & 65.8 & 79.9 & 76.8 & 20.1 & 1.62 \\ \hline
YOLOv10n & 55.2 & 74.9 & 6.7 & 2.3 & 1.71 \\ \hline
YOLOv10s & 61.5 & 78.2 & 21.6 & 7.2 & 1.65 \\ \hline
YOLOv10m & 64.8 & 80.0 & 59.1 & 15.4 & 1.67 \\ \hline
\end{tabular}%
}
\end{table}

In total, we train 60 models, comprising 6 clean models—one for each YOLOv9 and YOLOv10 variant trained on the clean dataset—and 54 poisoned models, where each YOLO variant is trained on 9 different poisoning ratios. In this way, we aim to investigate the impact of the extent of poisoning and the complexity of the detectors on the success of the ShrinkBox attack. For evaluation, we employ the standard mAP@0.5:0.95 metric. To evaluate infected detectors specifically, we compare mAP\(_{\text{clean}}\), mAP\(_{\text{pois}}\), AP\(_{\text{clean}}\), and AP\(_{\text{pois}}\) for each infected detector, where AP corresponds to the attacked Car class and the \textbf{clean} and \textbf{pois} metrics correspond to inference on clean samples and poisoned samples respectively. Table~\ref{bases} shows the results of the 6 baseline detectors trained on clean data in terms of their object detection accuracy, efficiency, and distance estimation accuracy, in MAE, when combined with DECADE. 

We present the results of the performance of the infected models at different poisoning ratios during training in Fig.~\ref{evalDet}. Additionally, we include the ASR scores at the most relaxed similarity threshold of 0.5 in the figure. Firstly, we find that larger models, in terms of scale, indeed outperform their smaller counterparts in both mAP and AP over both the clean and poisoned instances. Most importantly, however, the larger models are more prone to the attack. For instance, the YOLOv9 t, s, and m achieve ASR scores of 81.5\%, 88.3\%, and 91.9\% respectively at only a 2\% poisoning ratio for instance. Furthermore, when trained on the 1\% ratio, the YOLOv10n only scores an ASR of 31.3\%, while the YOLOv10m scores 87.3\%. We attribute this to the larger models’ greater capability to learn the diverse associations, including the poisoning, present in the dataset. We also observe that the ASR tends to improve with a greater poisoning ratio. This is expected since more poisoned samples allows the models to better learn the association between the triggers and the shrunken boxes. 

Secondly, our hypothesis that the difference between mAP\(_{\text{clean}}\) and mAP\(_{\text{pois}}\) is negligible holds only if the poisoning ratio is at least 5\%. This can be explained by investigating the AP of the Car class which was attacked specifically. The difference in AP\(_{\text{clean}}\) and AP\(_{\text{pois}}\) is $\le$ 1.5\% in detectors infected at the 10\% and 20\% ratios. Interestingly, at lower poisoning ratios, AP\(_{\text{pois}}\) is much lower than AP\(_{\text{clean}}\). However, it starts to exceed AP\(_{\text{clean}}\) at ratios greater than 10-20\%. We believe that this is because at low ratios, the models cannot learn the poisoned association well due to not enough poisoned samples being available. The opposite is true in the case of high ratios, where the models start to favor the poisoned over the clean associations due to the greater abundance of poisoned instances in the dataset. 

Furthermore, we find that the differences between AP\(_{\text{clean}}\) and AP\(_{\text{pois}}\) start to decrease as the model’s complexity increases. For instance, we observe the greatest differences between these metrics at the 1\% poisoning ratio in the YOLOv9 t, s, and m models as 15.3\%, 8.2\%, and 6.8\% respectively, where they clearly decrease as the scale grows. A similar trend is observed when the AP\(_{\text{pois}}\) is greater than AP\(_{\text{clean}}\) by the largest difference at the poisoning ratio of 75\%. These trends also apply to the YOLOv10 models. We believe that they can be explained by the greater learning capability of larger models. In this way, we find that the larger models are the easiest to attack with ShrinkBox while also ensuring the quality of performance in the clean samples. Furthermore, regardless of the model complexity, we recommend the poisoning ratio to be between 10-20\% in order to maximize the ASR while also minimizing the differences between AP\(_{\text{clean}}\) and AP\(_{\text{pois}}\).

Lastly, even when a predicted bounding box is considered as successfully attacked, its deviation from the dimensions of the shrunken ground-truth box might vary greatly. Thus, an in-depth analysis of the ASR over different similarity thresholds is required to determine how closely the predictions of infected detectors on images containing the trigger align with the poisoned instances, as opposed to the clean ones. Fig.~\ref{evalASR} visualizes our results where we set similarity threshold values inspired by the IoU intervals in mAP evaluation. We observe that ShrinkBox benefits from higher poisoning ratios in the training set as ASR scores for every model across the thresholds increase with an increase in the poisoning ratio. Interestingly, the differences between the ASR scores of infected models across size scales continue to decrease, with the scores becoming almost identical at the 20\% poisoning ratio for the YOLOv9 models. Furthermore, ASR scores tend to decrease as we increase the strictness of similarity-based matching with higher thresholds. However, the ASR at these higher thresholds better reflects the ability of an infected model to predict shrunken boxes that align with our critical offset distance of 5m.

\subsection{Evaluation of DECADE with Infected YOLO Models}

Finally, we evaluate the impact of all the infected YOLO models on detection-wise distance estimation with DECADE\textcolor{green}{~\cite{decade}} over the 1701 potentially poisoned instances in the validation set using the MAE metric. The baseline MAE values, with the clean YOLO models, are presented in the final column of Table~\ref{bases}. For the infected models, we compare the distance predictions with the corresponding clean ground-truth distances and compute MAE\(_{\text{clean}}\) when the trigger is absent and MAE\(_{\text{pois}}\) when the trigger is present in the images. Note that our poisoning strategy shrinks bounding boxes to project them to a distance that is further than the ground-truth by a critical offset. Since our critical offset is 5m, in the case of a perfectly infected detector, we would expect MAE\(_{\text{pois}}\) to be at least 5m. With that said, we present our results in Fig.~\ref{evalDECADE}.

Firstly, we observe that MAE\(_{\text{clean}}\) remains close to the baseline values when DECADE is combined with clean detectors. This is ideal since we expect normal behavior on clean samples. On the other hand, as ASR starts to increase, so does MAE\(_{\text{pois}}\). This is because having more successfully attacked samples within the poisoned samples degrades the MAE by larger margins. In line with this observation, we find that the lowest MAE\(_{\text{pois}}\) occurs when the ASR is also the lowest at the 1\% poisoning ratio, with YOLOv9t yielding 3.63m at 63.2\% ASR and YOLOv10n yielding 2.56m at 31.8\% ASR. Similarly, the highest MAE\(_{\text{pois}}\) of 5.5m occurs when the ASR is second to the highest at 97.2\% in the YOLOv9m model trained with a 50\% poisoning ratio. Since high ASRs are recorded in detectors infected at higher poisoning ratios, MAE\(_{\text{pois}}\) values also increase with higher poisoning ratios. Overall, we find that ASR scores $\ge$ 95 leads to at least an MAE\(_{\text{pois}}\) of 5m. Lastly, considering our recommended poisoning ratios in the previous subsection, we find that the 10-20\% poisoning ratios indeed lead to the MAE\(_{\text{pois}}\) of 5m, except for the YOLOv9s, YOLOv10n, and YOLOv10m where these models yield 4.94m, 4.98m, and 4.87m respectively– all close to the critical offset.


\section{Conclusion}

In this work, we introduce ShrinkBox, a novel backdoor attack targeting object detection models in the safety-critical application of collision avoidance in ML-ADAS. Unlike existing attacks that introduce conspicuous modifications to object annotations, ShrinkBox subtly shrinks bounding boxes in poisoned instances, making the attack nearly undetectable through manual inspection or standard evaluation metrics like mAP. We further propose a novel ASR metric to effectively measure the impact of ShrinkBox and demonstrate a 96\% ASR is achievable with only a 4\% poisoning ratio in the training set. While mAP remains unaffected, we showed that ShrinkBox significantly degrades downstream distance estimation in models like DECADE, where key features extracted from the detected objects are relied upon, increasing the MAE by 3.1x, eventually reaching an error of 5 meters, on poisoned instances. Thus, by shrinking object bounding boxes such that they correspond to distances that are more than the average length of a car (4.5m) farther than the ground-truth, ShrinkBox manipulates the perception of object proximity, leading to potential crashes as collision warnings may be delayed or not generated at all. Our findings highlight the severe risks posed by imperceptible manipulations in object detection, underscoring the need for more robust defenses against backdoor attacks in ML-ADAS to safeguard autonomous systems from such stealthy vulnerabilities that may result in tragic consequences.

\section*{Acknowledgment}
This research was partially funded by Technology Innovation Institute (TII) under the "CASTLE: Cross-Layer Security for Machine Learning Systems IoT" project.

\bibliographystyle{IEEEtran} 
\bibliography{main} 

\end{document}